\pdfoutput=1

\documentclass[10pt,twocolumn,letterpaper]{article}

\usepackage[pagenumbers]{cvpr}

\usepackage{algorithm}
\usepackage{algpseudocode}
\usepackage{multirow}

\usepackage{placeins} %

\usepackage{xspace}
\newcommand{\plain}{Plain\xspace}
\newcommand{\cacheonly}{Cache-only\xspace}
\newcommand{\chainonly}{Chain-only\xspace}

\newif\ifsuppincluded \suppincludedtrue
\newcommand{\suppref}[1]{\cref{#1}}
\AtBeginDocument{\crefname{appendix}{Appendix}{Appendices}\crefname{subappendix}{Appendix}{Appendices}\crefname{subsubappendix}{Appendix}{Appendices}}

\definecolor{linkblue}{rgb}{0.21,0.49,0.74}
\usepackage[pagebackref,breaklinks,colorlinks,allcolors=linkblue]{hyperref}

\title{Memento: Long-Horizon Reasoning in Video Generation \\ by
Writing States into Pixels}

\author{Minghao Tian$^{1}$ \quad Yifei Wang$^{1}$ \quad Feng Wang$^{2}$ \quad Chen Wei$^{1}$ \quad Alan Yuille$^{2}$\\[4pt]
$^{1}$Rice University \qquad $^{2}$Johns Hopkins University\vspace*{-10pt}}

\begin{document}
\input{figures/teaser}
\begin{abstract}
Video generation models can solve reasoning problems in pixels, but their success collapses as the reasoning horizon---the number of dependent steps a problem takes---grows. A single generation denoises all frames jointly, so no intermediate result is fixed before the steps that depend on it are computed. We propose Memento. Its {visible cache} renders the hidden state of the process---a running sum, the pose of an object, the velocity of a ball---into the frame. Its {chain of passes} generates the video in short passes, each conditioned only on the last frame of the previous pass and the unchanged instruction, so every pass starts from a frame that shows everything computed so far and takes only the next step. Memento requires no architectural change, only image-to-video fine-tuning on videos rendered with a task-specific cache; at inference the model receives just the first frame and the instruction, and writes and reads the cache itself. On six procedurally generated tasks in arithmetic, spatial and physical reasoning, with the horizon controlled, Memento raises average success at horizon 4 from 0.16 for standard single-pass generation to 0.66 on Wan2.2-TI2V-5B, and retains 0.54 at horizon 8, where the single pass falls to 0.07; the gains at horizon 4 carry over to LongCat-Video. Gains are largest where a plain frame hides the whole state and smallest on the physical tasks, where it hides only the velocity. With the training instances and the number of frames held fixed, the cache without the chain and the chain without the cache both fall short of Memento on every task and on both base models tested.
\end{abstract}
\section{Introduction}
\label{sec:intro}
In Christopher Nolan's \emph{Memento}~\cite{memento_film}, Leonard Shelby cannot form new memories, yet he pursues a goal far longer than anything he can hold in mind. He writes his state down---on Polaroids, on notes, on his own skin---and works in short episodes, each of which begins by reading what the previous one left. He cannot remember, but he can read. A video generation model asked to reason through a long problem has Leonard's limit without his method: it must work everything out at once, with nothing written down.

Video generation models are increasingly used to simulate, plan and solve problems in pixels~\cite{genie, unipi, vbvr, wiedemer, unisim, videolang, thinkingvideo}. In each of these roles the model must get a long chain of dependent steps right: each step is determined by the one before it, and a single wrong state invalidates everything after it. We find that the two image-to-video models we fine-tune cannot. Fine-tuned to add numbers shown in its first frame, a model adds two correctly every time, but five in turn only 0.07 of the time, and nine never. On a waypoint maze the ball circles back along its own path and returns to waypoints it has already visited (\cref{fig:teaser}), or cuts through walls. We measure this on procedurally generated tasks in which the number of dependent steps, the \emph{reasoning horizon} $N$, is a controlled variable and success is checked by rule, so that failures can be traced to the horizon rather than to the difficulty of any one step taken alone.

We attribute this to how a clip is generated. A diffusion model denoises all frames together: every denoising step updates the whole clip, and attention couples each frame to every other. The frames that should show an intermediate result are therefore formed at the same time as the frames that should build on it, and no result is fixed before the steps that depend on it are computed; consistent with this, analyses find that reasoning in these models unfolds along the denoising steps rather than across frames~\cite{cos}. Language models escape this limit through chain of thought, where each token is fixed once written and is read by every step that follows, so that serial computation grows with the length of the chain~\cite{li, merrill, wei}. A video model lacks both halves of this mechanism. Nothing is fixed until the generation ends. And even a finished frame would not be enough to continue from: a plain frame shows at most where the process is and what has been output so far, not the state the next step depends on---a running sum and its carry, the pose of a die, the velocity of a ball at that instant (\cref{fig:cache}).

We propose Memento, which supplies both halves the way Leonard does: it writes the state down, and works in short episodes that each begin by reading it. The \emph{visible cache} writes the hidden state of the process into the pixels, and the \emph{chain of passes} generates the video as a sequence of short passes, each conditioned only on the last frame of the previous pass and the unchanged instruction (\cref{fig:teaser}). Each pass starts from a frame in which everything computed so far is visible, and only has to take the next step; the number of passes becomes the axis along which reasoning grows, as the length of the chain does for a language model. Memento changes nothing in the architecture: the cache is designed per task and rendered into the training videos, so image-to-video fine-tuning teaches the model to write it and to read it through its existing image conditioning. At inference the model receives only the first frame and the instruction, and because the cache is an image, what it has computed can be inspected directly at every handoff (\cref{fig:qual}).

We evaluate on six tasks in three domains, judged either on the final frame or on the whole trajectory. At $N{=}4$ on Wan2.2-TI2V-5B, Memento raises average success from 0.16 for standard single-pass generation to 0.66. The gain is largest where a plain frame hides the state (0.07 to 0.93 on chain addition) and smallest on the physical tasks, where it hides only the velocity: bouncing under gravity improves from 0.29 to 0.42 and remains far from solved. The gains carry over to LongCat-Video (\cref{tab:main}). At $N{=}8$ the single pass averages 0.07, and Memento 0.54 (\cref{fig:scaling}, \textit{top}).

Prior work on video reasoning has pursued each half on its own. Chain-of-frames reasoning~\cite{wiedemer} renders intermediate results, but within a single clip, where nothing is fixed before it is read; chaining generations from their final frames~\cite{newman}, the simplest form of chunk-by-chunk long-video generation~\cite{diffusionforcing, causvid}, fixes each chunk but writes no state into the frame it hands over. We test both in a controlled $2{\times}2$ comparison with the same training instances and the same total number of frames (\cref{tab:main,fig:qual}): the cache within a single pass averages only 0.18 at $N{=}4$, and the chain without the cache reaches 0.00 on chain addition. The gain of Memento thus comes from neither the added supervision nor the added compute alone, but from a state that is written and then fixed before it is read (\cref{sec:ablations}).

In summary, our contributions are:
\begin{itemize}
  \item We show that the success of single-pass image-to-video generation collapses as the reasoning horizon grows, extending what is known for maze path length~\cite{newman} to six generated tasks in which the horizon is controlled.
  \item We propose Memento, which writes the state of the process into a frame as a visible cache and hands that frame to a new pass, making the number of passes an axis of serial computation with no architectural change.
  \item We separate the two parts in a controlled $2{\times}2$ study and show that neither the visible cache within a single pass nor the chain of passes without it suffices on its own.
\end{itemize}
\section{Method}
\label{sec:method}

\subsection{Preliminaries and problem formulation}
\label{sec:setting}

\paragraph{Image-to-video diffusion.}
Current video generation models are predominantly latent diffusion transformers trained with flow matching~\cite{flowmatching, rectifiedflow, sd3, moviegen, hunyuanvideo, wan, longcat}, including both base models we use. A clip is encoded into a latent $\mathbf{z}$ that covers all of its frames. Given Gaussian noise $\boldsymbol{\epsilon}$ and a noise level $\sigma \in [0,1]$, the model $v_\theta$ is trained on the interpolation $\mathbf{z}_\sigma = (1-\sigma)\mathbf{z} + \sigma\boldsymbol{\epsilon}$ to predict the velocity $\boldsymbol{\epsilon} - \mathbf{z}$, conditioned on an image and a text prompt, collectively $\mathbf{c}$. Generation integrates this velocity field from pure noise to data over $S$ denoising steps $1 = \sigma_0 > \cdots > \sigma_S = 0$:
\begin{equation}
    \mathbf{z}_{\sigma_{i+1}} = \mathbf{z}_{\sigma_i} + (\sigma_{i+1} - \sigma_i)\, v_\theta\!\left(\mathbf{z}_{\sigma_i}, \sigma_i, \mathbf{c}\right).
    \label{eq:fm}
\end{equation}
Two properties of \cref{eq:fm} matter. Every denoising step updates the latent of the entire clip at once, so no frame is finished before the others. And the transformer attends across the tokens of all frames, so the update of each frame depends on the estimate of every frame, earlier and later.

\paragraph{Task formulation.}
A task instance is specified by an initial frame $\mathbf{x}_0$ that depicts a problem, together with a text instruction $\tau$ that is fixed for the instance. A video diffusion model produces a clip $\mathbf{x}_{0:T-1}$ of $T$ frames that starts from $\mathbf{x}_0$; in \emph{final-state} tasks only the last frame $\mathbf{x}_{T-1}$ is scored, in \emph{trajectory} tasks the whole clip. In the standard setting the entire clip is produced by a single denoising run,
\begin{equation}
    \mathbf{x}_{0:T-1} \sim p_\theta\!\left(\,\cdot \mid \mathbf{x}_0, \tau\right),
    \label{eq:single}
\end{equation}
which we refer to as a \emph{single pass}: in terms of \cref{eq:fm}, $\mathbf{c} = (\mathbf{x}_0, \tau)$ and the latent spans all $T$ frames.

The tasks we study share one structure: the output is determined by a sequence of $N$ dependent steps that take the process from the initial state $s_0$, given by $\mathbf{x}_0$, through states $s_1 \to s_2 \to \cdots \to s_N$, each computed from the previous one---a running sum and its carry, the pose of a die after each instructed move, the velocity of a ball after each interval of its trajectory. We call $N$ the reasoning horizon, or simply the horizon. In both kinds of task, the states are fixed by the problem, and the video only renders them. Memento replaces the single run of \cref{eq:single} with a chain of passes that communicate through their boundary frames, so that no pass has to span more than one step.

\subsection{Memento}
\label{sec:vsp}

\begin{figure*}[t]
  \centering  \includegraphics[width=1\linewidth]{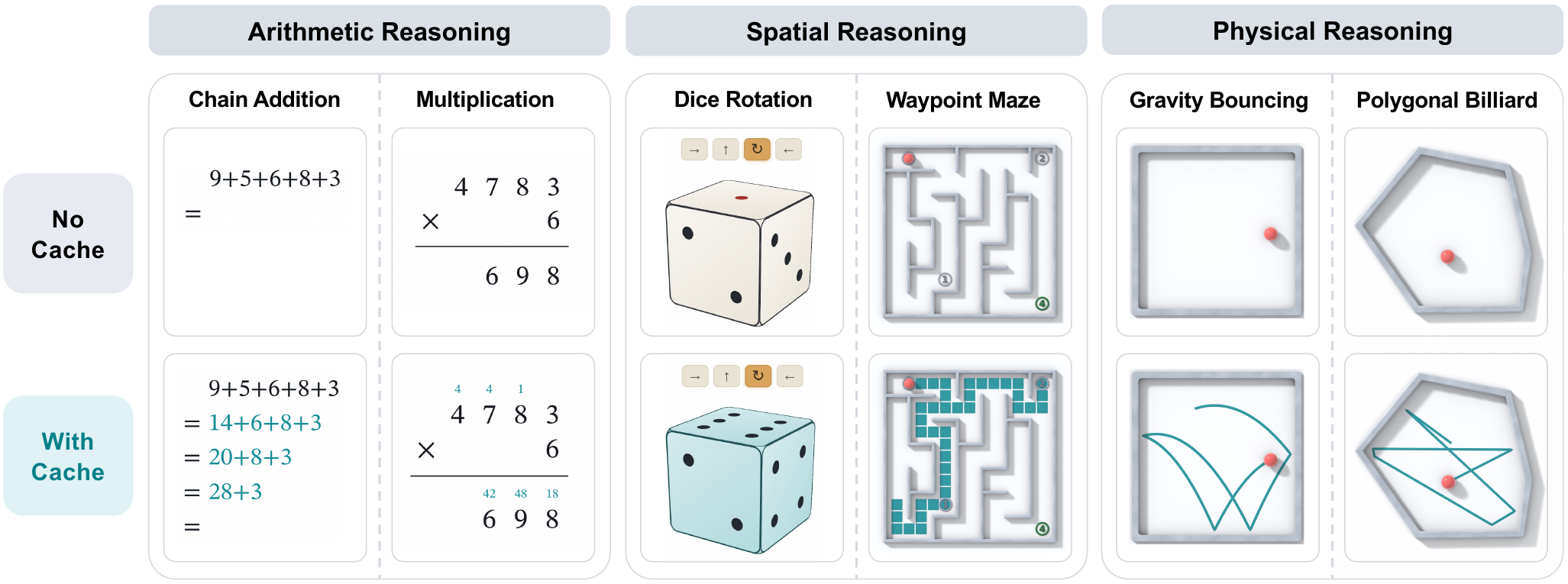}
\caption{\textbf{The visible cache across the six tasks.} For each task at $N{=}4$, the boundary frame after the third pass, without and with the visible cache. The cache, in teal, adds the state the next step depends on: the running sum (\textsc{Add}), the partial products and carries (\textsc{Mul}), the die in its current pose (\textsc{Dice}), the visited cells (\textsc{Maze}), and the ball's trail, whose shape reveals its velocity (\textsc{Gravity}, \textsc{Billiard}).}
  \label{fig:cache}
\end{figure*}

A single pass fixes no state before the states that depend on it are computed (\cref{sec:setting}), whereas in chain of thought each token is fixed once written and read by every later step. Memento gives a frame both properties, one component each: the visible cache writes the state into the frame, so that the next step can be computed from it, and the chain of passes starts each step from a frame an earlier pass finished.

\paragraph{The visible cache.}
The next step should be computable from the frame alone: $s_{k+1}$ should follow from what the frame shows in a single step of the process, without recomputing $s_1, \dots, s_{k-1}$. In other words, the process should be Markov in the frame. A plain rendering rarely meets this: it shows at most where the process is and what has been output so far, not the state the next step depends on, so the next step would have to redo the earlier ones, even though the problem, still legible in the frame, determines them in principle. Memento therefore writes what the remaining steps depend on \emph{into the frame}. We call this record the \emph{visible cache}. The name is by analogy with the KV cache of an autoregressive language model, which stores what earlier steps have computed so that later steps need not recompute it. The visible cache does the same for the chain of passes, but in pixels, the only channel that a pass conditioned on a single frame can read. Its content is dictated by the hidden state of the process, whatever form that takes (\cref{sec:cache}).

\paragraph{The chain of passes.}
Each state should also be fixed before the steps that read it are computed, which a single pass cannot provide. We therefore divide the video into $K$ chunks of $L$ frames each (\cref{fig:teaser}), with $T = K(L-1)+1$ so that consecutive chunks share exactly one frame and the chain generates exactly as many frames as a single pass. Chunk $k$ covers frames $(k-1)(L-1)$ to $k(L-1)$ of the video and is produced by a separate pass of the same model, conditioned only on the last frame of chunk $k-1$, where the visible cache is written, and the instruction:
\begin{equation}
    \mathbf{x}^{(k)}_{0:L-1} \sim p_\theta\!\left(\,\cdot \mid \mathbf{x}^{(k-1)}_{L-1}, \tau\right),
    \qquad \mathbf{x}^{(0)}_{L-1} \equiv \mathbf{x}_0 .
    \label{eq:chain}
\end{equation}
The first frame of each chunk is its conditioning frame, and the full video is the concatenation of the chunks with shared frames deduplicated. No pass sees any frame other than its conditioning frame; in particular, $\mathbf{x}_0$ is not re-supplied, so the problem itself must remain legible in every boundary frame. The instruction $\tau$ is identical for every pass and no pass index is given, so no information enters the computation after $\mathbf{x}_0$, and a pass can tell which step to take only from the frame it starts from. With $K{=}N$, as in our main experiments, each pass takes one step, starting from a frame in which the previous state is already fixed, so a video of the same total length is computed as $K$ steps in series rather than all at once; with fewer passes, each takes several (\cref{sec:ablations}). All chunks have the same length, and the videos are rendered so that every step takes the same number of frames; a chunk boundary therefore always falls where a step is complete and its state has been written. In continuous processes the trail makes any instant a valid handoff.

\paragraph{Training and inference.}
Memento is trained on the same instances as the single-pass baseline, with only two differences: the ground-truth videos are rendered with the cache, and each is cut into $K$ clips of equal length $L$, each conditioned on its own first frame and $\tau$. The architecture and the objective are unchanged: the model learns to write the cache like any other content of the clip, and reads it through its image conditioning. At inference only $\mathbf{x}_0$ and $\tau$ are given. The $K$ passes of \cref{eq:chain} run in sequence, each an ordinary image-to-video generation that starts from the last frame generated by the previous pass, so the cache a pass reads is the one the model itself wrote; no ground-truth frame or cache is supplied at any point. $K$ is fixed in advance. 

\subsection{Designing the visible cache}
\label{sec:cache}
Two principles govern what goes into the boundary frame. It must be \emph{complete}: every quantity the next step depends on is present, including how far the process has come, so that the process is Markov in the frame (\cref{sec:vsp}). It should be \emph{incremental}: a step changes only the state it has produced, and whatever else earlier steps wrote stays in the frame rather than being redrawn. What this requires differs by task (\cref{fig:cache}). Where the state has no visual form of its own---a running sum and its carry---the cache is an explicit annotation written into the frame. Where the state is an object's configuration---the pose of a die---the cache is the object shown in that state. Where the frame shows position but hides history or velocity---a maze, a bouncing ball---the cache is a trail whose shape recovers what the frame hides. No trail exists at $t{=}0$, so in the physical tasks the initial frame carries an indicator of the initial velocity; it is part of the problem and is present with or without the cache.

\section{Experiments}
\label{sec:experiments}

\begin{table*}[t]
  \caption{\textbf{Success rate at $N{=}K{=}4$} across tasks and two base models; \cref{fig:scaling} varies $N$. All four schemes generate the same $T{=}193$ frames from the same training instances, the chained ones with one step per pass of $49$ frames. Columns follow the $2{\times}2$ design and success criteria of \cref{sec:setup}. $^\dagger$Trajectory task, judged on the whole clip; the others are final-state tasks, judged on the last frame. Best per model in bold.}
  \label{tab:main}
  \centering
  \setlength{\tabcolsep}{3.8pt}
  \begin{tabular}{@{}llcccc@{\hskip 1.5em}cccc@{}}
    \toprule
     & & \multicolumn{4}{c}{Wan2.2-TI2V-5B~\cite{wan}} & \multicolumn{4}{c}{LongCat-Video~\cite{longcat}} \\
    \cmidrule(lr){3-6} \cmidrule(l){7-10}
    Domain & Task & \plain & \cacheonly & \chainonly & Memento & \plain & \cacheonly & \chainonly & Memento \\
    \midrule
    \multicolumn{2}{l}{\emph{Visible cache}}   & -- & \checkmark & -- & \checkmark & -- & \checkmark & -- & \checkmark \\
    \multicolumn{2}{l}{\emph{Chain of passes}} & -- & -- & \checkmark & \checkmark & -- & -- & \checkmark & \checkmark \\
    \midrule
    \multirow{2}{*}{Arithmetic}
      & Chain addition                         & 0.07 & 0.04 & 0.00 & \textbf{0.93} & 0.09 & 0.09 & 0.00 & \textbf{0.95} \\
      & Multiplication                         & 0.13 & 0.12 & 0.14 & \textbf{0.73} & 0.21 & 0.17 & 0.29 & \textbf{0.82} \\
    \midrule
    \multirow{2}{*}{Spatial}
      & Dice rotation                          & 0.00 & 0.00 & 0.00 & \textbf{0.72} & 0.00 & 0.00 & 0.00 & \textbf{0.86} \\
      & Waypoint maze$^\dagger$                & 0.37 & 0.60 & 0.46 & \textbf{0.89} & 0.64 & 0.83 & 0.69 & \textbf{0.98} \\
    \midrule
    \multirow{2}{*}{Physical}
      & Gravity bouncing$^\dagger$             & 0.29 & 0.22 & 0.09 & \textbf{0.42} & 0.28 & 0.23 & 0.18 & \textbf{0.43} \\
      & Polygonal billiard$^\dagger$           & 0.07 & 0.09 & 0.01 & \textbf{0.29} & 0.02 & 0.04 & 0.00 & \textbf{0.15} \\
    \midrule
    \multicolumn{2}{l}{Average}                              & 0.16 & 0.18 & 0.12 & \textbf{0.66} & 0.21 & 0.23 & 0.19 & \textbf{0.70} \\
    \bottomrule
  \end{tabular}
\end{table*}

\subsection{Setup}
\label{sec:setup}

\paragraph{Tasks and success criteria.}
We study six procedurally generated tasks in three domains (\cref{fig:cache}); each instance is fully specified by its initial frame, and success is decided by a rule-based scorer that is the same for all four schemes. \emph{Arithmetic.} \textsc{Add} adds $N{+}1$ one-digit terms left to right, one addition per step. \textsc{Mul} multiplies an $N$-digit number by a nonzero one-digit number, one digit position per step from the units digit up, each position depending on the carry from the one before. Both succeed if the final result is exactly correct. \emph{Spatial.} \textsc{Dice} asks for the pose of a die after $N$ rotation instructions, one per step, and succeeds if the final pose is correct. A waypoint maze (\textsc{Maze}) asks a ball to visit $N$ waypoints in a prescribed order in a grid maze, and succeeds if a single ball moves continuously, never crosses a wall and visits all waypoints in order; any valid path counts. \emph{Physical.} In \textsc{Gravity} a ball is launched into a box and bounces under gravity; in a polygonal billiard (\textsc{Billiard}) the ball moves at constant speed inside a random convex hexagon, reflecting elastically off its edges. Both have a unique ground-truth trajectory, and an instance succeeds if a single ball is within two ball radii of its ground-truth position in at least $80\%$ of the frames (track@2R), which penalizes a right path at the wrong speed. The first three are final-state tasks and the last three trajectory tasks (\cref{sec:setting}); judging the latter on the whole clip~\cite{vbvr, vbvrpro, viper} means that any error at a handoff counts against Memento. Rendering and scorer details are in \suppref{sec:supp_tasks,sec:supp_metrics}.

The horizon $N$ is the number of additions, digit positions, instructions or waypoints. In the physical tasks the steps are not discrete operations: the state evolves continuously, each instant depending on the one before, and $N$ is the simulated time span in units of a fixed interval. Every clip has $T{=}193$ frames at every $N$, so a larger $N$ packs more dependent steps into the same clip, $192/N$ frames each; in the physical tasks this shows as faster motion per frame and more bounces. At a given $N$ all four schemes render the same steps at the same rate. Since the single pass succeeds at $N{=}1$ with the same clip length and compute, its decline with $N$ is not due to clip length; wherever Memento succeeds with the same frames per step, it is not due to too few frames per step either. Clips at different $N$ differ only in the number of dependent steps they contain; in the physical tasks, motion per frame grows with $N$ as well.

A separate model is trained for each task and horizon. Test instances are held out from training, $200$ per task at $N{\le}2$ and $500$ at $N{\ge}4$ (split sizes in \suppref{tab:supp_splits}); training uses $3000$ instances, or all remaining ones where the instance space is smaller (\textsc{Add}, \textsc{Mul} and \textsc{Dice} at $N{\le}2$). Instances are held out, not steps: in \textsc{Add}, \textsc{Mul} and \textsc{Dice} the same single steps recur across instances, so a one-step pass is tested on new sequences of steps it has seen, which is what decomposition buys. The exception is \textsc{Add}, \textsc{Mul} and \textsc{Dice} at $N{=}1$, which have only $81$--$144$ instances and are trained, validated and evaluated on all of them; these cells measure whether a single step can be fit, not whether it generalizes. We report success rate as the fraction of evaluated instances that succeed.

\begin{figure*}[t]
  \centering
  \includegraphics[width=\linewidth]{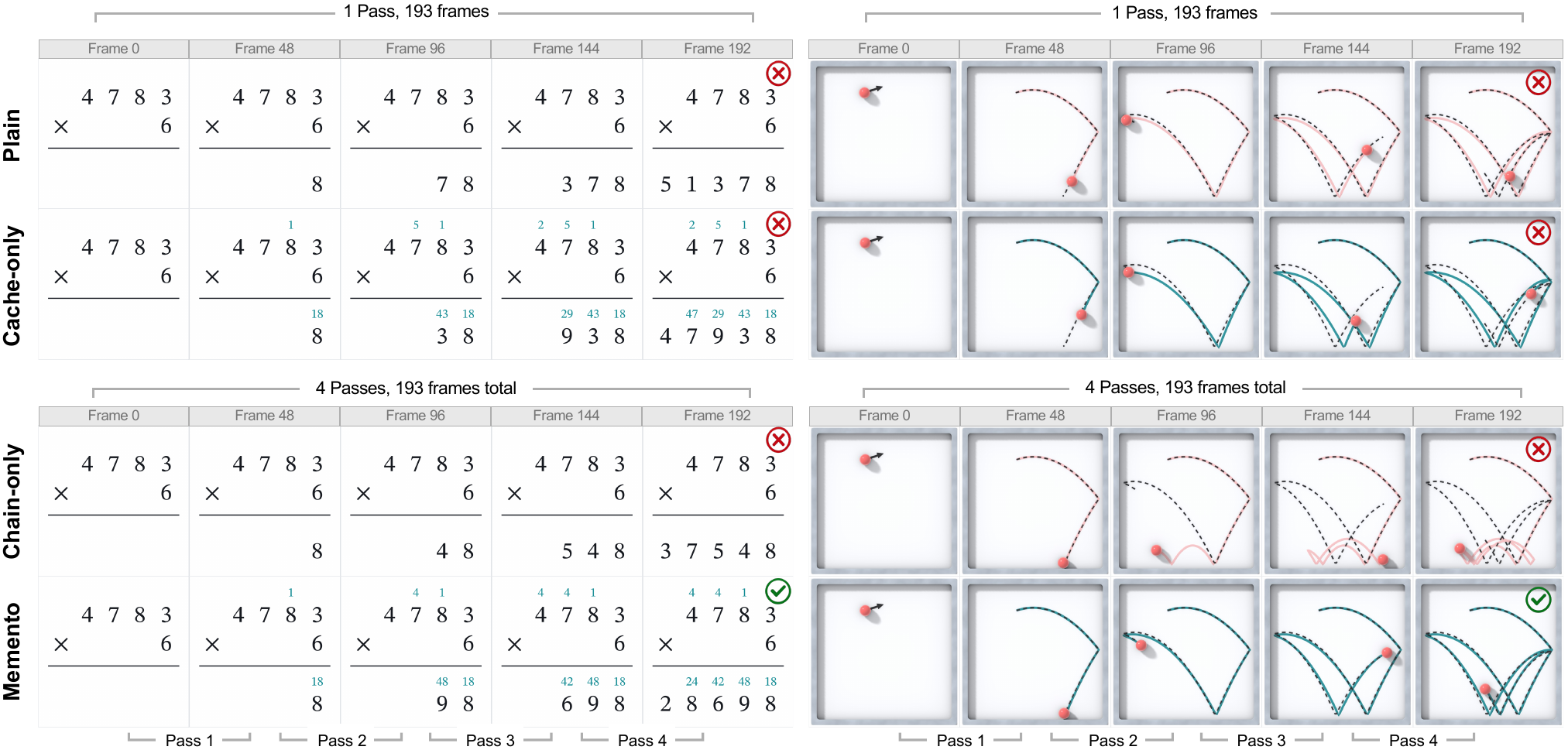}
  \caption{\textbf{Qualitative comparison of the four schemes} at $N{=}4$ on \textsc{Mul} and \textsc{Gravity}. Columns are the pass boundaries of the chained schemes (bottom) and the same frames of the single-pass schemes (\emph{top}). Teal marks the visible cache, generated by the model: partial products and carries in \textsc{Mul}, the trail in \textsc{Gravity}. In \textsc{Gravity}, the dashed line (ground truth) and, in the two rows without a cache, the light red path (the track of the generated ball) are overlaid for display only. \cacheonly writes wrong values into the cache; \chainonly loses the carry and the velocity across handoffs; only Memento stays correct throughout.}
  \label{fig:qual}
\end{figure*}

\paragraph{Baselines: the {\boldmath$2{\times}2$} design.}
\label{sec:design}
Toggling the cache and the chain gives four generation schemes, listed in \cref{tab:main}, all of which generate the same number of frames $T$; \cref{fig:qual} shows them on two instances. \plain, a single pass without the cache, is the standard image-to-video setting. \cacheonly, a single pass with the cache, follows work that renders intermediate steps within one clip~\cite{wiedemer, abbe}: it writes intermediate results, but each is formed together with the steps that should read it, so it is never handed off. \chainonly, a chain of passes without the cache, follows work that chains generations from their final frames~\cite{newman}: it fixes each boundary frame before the next pass but writes no state into it. The frame carries only what a plain rendering shows---a highlight on the current operation in final-state tasks, the ball's position in trajectory tasks---so each pass knows where the process is but not what has been computed. \cacheonly versus Memento, trained on the same rendered videos, therefore isolates the chain, and \chainonly versus Memento, which share the same chain of passes, isolates the cache.

\begin{figure*}[t]
  \centering
  \includegraphics[width=\linewidth]{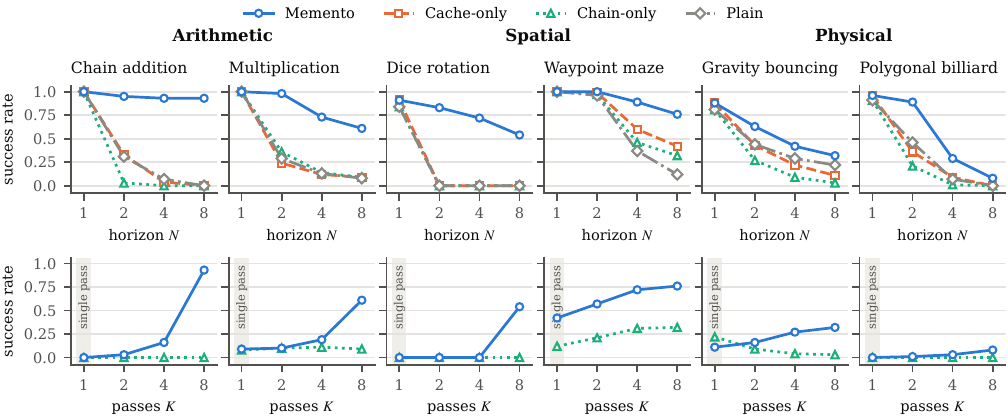}
\caption{\textbf{Scaling with the horizon and the number of passes} (Wan2.2-TI2V-5B). (\emph{top}) Success rate versus horizon $N$ with one step per pass ($K{=}N$), for all four schemes. At $N{=}1$ the single pass and the chain coincide, so \cacheonly equals Memento and \plain equals \chainonly. (\emph{bottom}) Success rate versus number of passes $K$ at fixed horizon $N{=}8$, so each pass takes $8/K$ steps. $K{=}1$ (shaded) is the single pass, where Memento reduces to \cacheonly and \chainonly to \plain.}
\vspace{-5pt}
  \label{fig:scaling}\label{fig:granularity}
\end{figure*}

\paragraph{Model and training.}
All models are LoRA fine-tunes of one of two base models: Wan2.2-TI2V-5B~\cite{wan}, a 5B-parameter text- and image-conditioned video diffusion model, and LongCat-Video~\cite{longcat}, a 13.6B-parameter model that supports text-to-video, image-to-video and video continuation in a single network. We train rank-32 adapters on the attention and feed-forward projections in bfloat16, at $512{\times}512$ for Wan2.2-TI2V-5B and $320{\times}320$ for LongCat-Video to reduce the training cost of the larger model, with a constant learning rate of $10^{-4}$ using DiffSynth~\cite{diffsynth}. Both models are trained with flow matching and attend across all frames (\cref{sec:setting}). Wan2.2-TI2V-5B is used for all experiments and LongCat-Video for the main results (\cref{tab:main}).

Every configuration of a task is trained on the same instances, rendered with the visible cache for \cacheonly and Memento and without it for \plain and \chainonly. Single-pass models see each video as one clip of $T$ frames; chained models see it sliced into $K$ clips of $L$ frames, each conditioned on its own first frame---the boundary frame it will receive at inference---and never attend across a chunk boundary. Since $K(L-1)+1 = T$, both are exposed to the same frames. Their per-step compute differs---chained models take more, smaller optimizer steps, single-pass models attend over $K$ times more frames per step---so rather than fix a budget we train every configuration for at least $16$ epochs and then until its success rate on $100$ validation instances, disjoint from the training and test sets, has not improved for $3$ consecutive epochs, and evaluate the checkpoint with the highest validation success once on the test set. At inference, every run uses the same sampler and noise schedule with $S{=}50$ denoising steps. Unless stated otherwise we use $N{=}4$ and $K{=}N$ passes, so that each pass handles one step; \cref{sec:scaling} varies $N$ and \cref{sec:ablations} varies $K$ at fixed $N$.

\subsection{Memento enables long-horizon reasoning}
\label{sec:results}

\paragraph{Memento solves what a single pass cannot.}
\Cref{tab:main} compares the four schemes at $N{=}4$. Memento is best on every task and on both base models, raising the average success rate of the single pass (\plain) from 0.16 to 0.66 on Wan2.2-TI2V-5B and from 0.21 to 0.70 on LongCat-Video. How much it gains depends on how much of the state a plain frame hides. Where none of it is visible---the running sum, the carry, the pose of the die---a single pass rarely succeeds (at most 0.21 on either model), while Memento reaches 0.72 to 0.95. On \textsc{Mul} with Wan2.2-TI2V-5B, even the few successes of \plain are nearly all instances with multiplier $1$, a ninth of the test set, whose product is a copy of the multiplicand. On \textsc{Maze} the frame shows where the ball is but does not mark which waypoints it has visited, and Memento raises success from 0.37 to 0.89. In the physical tasks only the velocity is hidden, and the gain is smallest (\textsc{Gravity} 0.29 to 0.42, \textsc{Billiard} 0.07 to 0.29).

\paragraph{Memento holds up as the horizon grows.}
\label{sec:scaling}
\Cref{fig:scaling} (\emph{top}) varies the horizon with one pass per step ($K{=}N$); its $N{=}4$ points match \cref{tab:main} for Wan2.2-TI2V-5B. At $N{=}1$ the average success rate is 0.93 for the single pass and 0.96 with the cache, where Memento is itself a single pass. A single step is therefore learnable in this setting, while failures at larger \(N\) are consistent with the difficulty of composing dependent steps. The single pass falls as $N$ grows, to 0.07 on average at $N{=}8$, and so do \cacheonly (0.10) and \chainonly (0.07); Memento retains 0.54. Memento declines most in the physical tasks, to 0.32 on \textsc{Gravity} and 0.08 on \textsc{Billiard} at $N{=}8$, where the trajectory grows with $N$ and small errors in position and velocity compound across passes; in the other four tasks it retains at least 0.54.

\subsection{Why it works}
\label{sec:ablations}

\paragraph{Both parts are necessary.}
Neither component suffices on its own, and how each falls short follows what the frame hides. In the final-state tasks, where the state must be computed before it can be written, neither helps: \cacheonly and \chainonly are at or below \plain on \textsc{Add} (0.04 and 0.00 against 0.07) and \textsc{Dice} (0.00 for all three), and about the same on \textsc{Mul} (0.12 and 0.14 against 0.13). On LongCat-Video the pattern is the same, except that \chainonly gains on \textsc{Mul} (0.29 against 0.21), still far below Memento (0.82). \cacheonly forms its intermediate results together with the steps that should read them; \chainonly starts each pass from a fixed frame, but one that shows only where the process is and the output so far. On \textsc{Maze}, where the position of the ball carries part of the state, each helps partially: \cacheonly raises \plain from 0.37 to 0.60 and \chainonly to 0.46, yet both remain well below Memento (0.89), since without the chain the trail is not fixed before it is read, and without the trail a pass must infer which waypoints have been visited from the position of the ball alone. In the physical tasks the two fall short for different reasons. \cacheonly does not help on \textsc{Gravity} (0.22 against 0.29) and changes little on \textsc{Billiard} (0.09 against 0.07): within a single pass the motion across frames already shows the velocity, so the trail adds little there. \chainonly falls below \plain (\textsc{Gravity} 0.09, \textsc{Billiard} 0.01), since each cut discards the velocity that a single frame does not show. The trail matters at a handoff, and the handoff needs the trail.

\Cref{fig:qual} shows one instance of \textsc{Mul} and one of \textsc{Gravity} under all four schemes. On \textsc{Mul}, \plain and \chainonly get the units digit, which needs no carry, and go wrong from the tens digit on; \cacheonly does write partial products and carries, but wrong ones from the second step; Memento writes every one correctly. On \textsc{Gravity}, the single pass does not fail for want of the dynamics: at $N{=}1$, where the clip spans a single interval, \plain succeeds on 0.81 of the instances (\cref{fig:scaling}, \emph{top}). Over four intervals, both single-pass schemes trace nearly the right path but fall behind it, and by the last frame the ball has covered only part of the ground-truth trajectory, consistent with small errors accumulating over a clip in which no state is fixed along the way. \chainonly fixes a frame at each handoff but not the velocity: after the first handoff the ball loses its motion and continues in small hops along the floor. Memento hands over a frame that shows both position and velocity, so each pass has to carry the ball through one interval only, and it stays on the ground-truth trajectory across all three handoffs.

\begin{table}[t]
  \caption{\textbf{More denoising steps do not close the gap.} Success rate at $N{=}K{=}4$ on Wan2.2-TI2V-5B. \plain uses the same checkpoint, sampler and initial noise at both $S$; at $S{=}200$ it matches the network evaluations of Memento ($K{=}4$ passes of $S{=}50$).}
  \label{tab:steps}
  \centering
  \small
  \begin{tabular}{@{}lccc@{}}
    \toprule
     & \multicolumn{2}{c}{\plain} & Memento \\
    \cmidrule(lr){2-3} \cmidrule(l){4-4}
    Task & $S{=}50$ & $S{=}200$ & $K{=}4$, $S{=}50$ \\
    \midrule
    Chain addition               & 0.07 & 0.07 & \textbf{0.93} \\
    Multiplication               & 0.13 & 0.13 & \textbf{0.73} \\
    Dice rotation                & 0.00 & 0.00 & \textbf{0.72} \\
    Waypoint maze      & 0.37 & 0.38 & \textbf{0.89} \\
    Gravity bouncing   & 0.29 & 0.31 & \textbf{0.42} \\
    Polygonal billiard & 0.07 & 0.07 & \textbf{0.29} \\
    \midrule
    Average                      & 0.16 & 0.16 & \textbf{0.66} \\
    \bottomrule
  \end{tabular}
\end{table}

\paragraph{The gain is not a matter of compute.}
A chain of $K$ passes takes $K$ times as many network evaluations as a single pass, each over about $1/K$ of the frames, so its total compute is similar in the frame-wise layers and about $K$ times lower in attention. \chainonly takes exactly as many network evaluations as Memento over the same frames, so the gap between them is due to the cache, not to the additional denoising steps. Since reasoning in video diffusion models has been found to unfold along the denoising steps rather than across frames~\cite{cos}, the single pass might instead close the gap with more denoising steps. It does not. Run for $KS{=}200$ steps, as many network evaluations as Memento at several times its compute, it changes by at most 0.02 on any task and averages 0.16, as with $S{=}50$, against 0.66 for Memento (\cref{tab:steps}). Additional steps refine the same jointly denoised clip: every step updates all frames together, so no intermediate result is fixed before the frames that depend on it are formed. What the single pass lacks is not the number of steps but a state that is fixed before it is read.

\paragraph{How many steps per pass?}
\Cref{fig:granularity} (\emph{bottom}) fixes the horizon at $N{=}8$ and varies the number of passes $K$, so that each pass takes $N/K$ steps and generates $192/K{+}1$ frames; at $K{=}1$, Memento reduces to \cacheonly and \chainonly to \plain. Success of Memento rises with $K$ on every task. In the final-state tasks nearly all of the gain comes at $K{=}8$, one step per pass: from 0.16 at $K{=}4$ to 0.93 on \textsc{Add}, from 0.19 to 0.61 on \textsc{Mul}, and from 0.00 to 0.54 on \textsc{Dice}. A pass that must take even two steps fails much as a single pass does, and the chain compounds its failures. \chainonly does not gain from more passes on these tasks (at most 0.11), since without the cache every pass must re-derive the state from a frame that does not show it. In \textsc{Gravity}, \chainonly falls from 0.22 at $K{=}1$ to 0.03 at $K{=}8$, since each cut discards the velocity, whereas Memento rises from 0.11 to 0.32, since the trail carries the velocity across the cut to the next pass.

\FloatBarrier
\section{Related work}
\label{sec:related}

\paragraph{Reasoning with video generation models.}
Video generation models have been shown to solve perception, physics and planning problems zero-shot, with the generated frames read as a chain of reasoning steps~\cite{wiedemer}, and benchmarks now evaluate this ability at scale~\cite{vbvr, vbvrpro, viper, mmecof, vreasonbench}, physical laws included~\cite{phyworld, physicsiq}. Fine-tuning strengthens it, on mazes~\cite{vrbench} and, with learnable reasoning tokens, across task families~\cite{opencof}. Analyses find that reasoning proceeds along the denoising steps rather than across frames~\cite{cos} and that a motion plan is committed early in the schedule~\cite{newman}. Most of this work studies reasoning within a single pass. We instead make the reasoning horizon a controlled variable and trace the failure of the single pass to the fact that it fixes no intermediate state. Closest in mechanism, Newman~\etal~\cite{newman} find that maze difficulty grows with path length and chain training-free generations, each conditioned on the final frame of the previous one. Such chaining is the simplest form of the chunk-by-chunk generation used for long videos~\cite{diffusionforcing, causvid, selfforcing, historyguided, framepack, magi1, rollingforcing, streamingt2v}, which aims at consistency over time rather than at computation: it carries forward what has been shown, not what has been computed. \chainonly is its single-frame form; more frames of history would keep motion visible across a cut, but add no state that the frames do not show. Consistent with Newman~\etal~\cite{newman}, \chainonly helps on our maze; where the state is not visible in the frame, it does not.

\paragraph{Writing intermediate state.}
Language models solve multi-step problems more reliably when they write out intermediate steps~\cite{nye, wei, kojima, anil}: a transformer of bounded depth performs a bounded amount of serial computation per forward pass, and chain of thought turns depth into length~\cite{feng, li, merrill, dziri}. Memento transfers this mechanism to video, with the boundary frame in the role of the emitted token and the visible cache as the state written into it. Vision-language models likewise reason better when they draw intermediate images~\cite{sketchpad, mvot, visualplanning}. In vision, the closest precedent is the chain-of-sketch of Lotfi~\etal~\cite{abbe}, who train vision transformers to emit intermediate sketches for globally defined tasks such as connectivity and mazes, and find that an inductive variant, in which each sketch is regressed from the previous one alone, generalizes better than predicting all sketches in parallel. Our setting differs in three respects. We study pretrained video diffusion models, where predicting all states in parallel fails for a specific reason: they are committed jointly across the denoising schedule~\cite{cos}. Each pass generates a clip rather than a frame, so the carried state can be dynamical, such as a velocity encoded by a trail, and the process itself can be the output. And we separate what is written from how it is handed over, showing in a controlled comparison that neither alone suffices. Their model also learns when to halt, whereas we fix $K$ in advance.
\section{Conclusion}
\label{sec:conclusion}
A single pass of a video diffusion model computes every step at once, so its success collapses as the reasoning horizon grows. Memento restores serial computation without architectural change: it writes the state into the pixels and hands it from one short pass to the next. On six tasks with Wan2.2-TI2V-5B, this raises average success from 0.16 to 0.66 at horizon 4, where either part alone averages at most 0.18, and from 0.07 to 0.54 at horizon 8. The cache is still designed by hand for each task and supervised with the ground-truth state, and one adapter is trained per task and horizon; learning what to write and transferring across horizons are left open. As with language models, what a video model can reason through appears to depend on whether it can write down what it has computed and read it back.

{
    \small
    \bibliographystyle{ieeenat_fullname}
    \bibliography{main}
}

\clearpage
\appendix
\setcounter{table}{0}\renewcommand{\thetable}{A\arabic{table}}
\setcounter{figure}{0}\renewcommand{\thefigure}{A\arabic{figure}}

\section{Implementation details}
\label{sec:suppl_impl}

The single-pass and chained models are trained with the same number of clips per optimizer step. Since a chained model sees each video as $K$ clips, it takes $K$ times as many optimizer steps per epoch, each on clips about $1/K$ as long (\cref{sec:setup}). Checkpoints are evaluated as saved after each epoch, without an exponential moving average of the weights.

\subsection{Tasks and data splits}
\label{sec:supp_tasks}

Every clip has $T{=}193$ frames for every task and horizon, so each of the $N$ steps spans $192/N$ frames; with $K{=}N$ passes, each pass generates $192/N{+}1$ frames and consecutive passes share their boundary frame. In \textsc{Gravity} and \textsc{Billiard} the horizon sets the simulated time span of the trajectory: the same $193$ frames cover $N$ intervals of fixed simulated duration, so a larger $N$ gives a longer trajectory with more bounces, and hence faster motion per frame; each step is one interval.

The length of $193$ frames is chosen so that the single pass and every chunking are valid inputs to the video VAE, without padding or trimming. Both base models encode video with a VAE that encodes the first frame on its own and compresses the remaining frames by a factor of four in time (Wan2.2-TI2V-5B with its own VAE, LongCat-Video in our setup with the Wan2.1 VAE), so a clip must have $4m{+}1$ frames. $193 = 4{\cdot}48{+}1$ frames give $49$ latent frames, and since $192$ is divisible by $2$, $4$ and $8$, the chunks of $192/K{+}1$ frames, \ie $97$, $49$ and $25$, are again of this form ($25$, $13$ and $7$ latent frames). The single pass and the chain of passes therefore cover exactly the same frames at every $K$. Frame height and width must be multiples of twice the spatial compression of the VAE, \ie $32$ for Wan2.2-TI2V-5B and $16$ for LongCat-Video; all tasks are rendered at $512{\times}512$ for Wan2.2-TI2V-5B and at $320{\times}320$ for LongCat-Video, which satisfies both.

All tasks are rendered at $24$\,fps, so every clip lasts about $8$\,s.

\paragraph{\textsc{Add}.} The terms are one-digit numbers from $0$ to $9$, written in one line, $a_0{+}a_1{+}\cdots{+}a_N$, with an equals sign on the line below (\cref{fig:cache}). With the cache, each step writes a new line below the previous one in which the first two terms are replaced by their sum, so that $9{+}5{+}6{+}8{+}3$ is followed by ${=}\,14{+}6{+}8{+}3$, then ${=}\,20{+}8{+}3$ and so on: the line written last always holds the running sum, followed by the terms still to be added. The new line fades in character by character from left to right. Without the cache, the result is written only in the last $20$ frames of the clip, which lie within the last step at every $N$.

\paragraph{\textsc{Mul}.} The problem is a column multiplication: the $N$-digit multiplicand, whose leading digit is nonzero, above the one-digit multiplier, and the result below the rule, aligned by digit position (\cref{fig:qual}). Step $k$ writes the $k$-th digit of the result, counted from the units digit, and the last step also writes the final carry as the leading digit. With the cache, the step also writes, in small digits, the product of the multiplicand digit and the multiplier directly above the result digit, and the carry directly above the next multiplicand digit, to which it is added in the next step. For $4783{\times}6$, the first step writes $18$ above the units digit $8$ of the result and the carry $1$ above the $8$ of the multiplicand.

\paragraph{\textsc{Dice}.} The initial pose is one of the $24$ orientations of the die, and each instruction is one of six rotations by $90^\circ$: \textsc{U} and \textsc{D} about one horizontal axis, \textsc{L} and \textsc{R} about the other, and \textsc{CW} and \textsc{CCW} about the vertical axis, which leave the top face in place; this gives $24{\cdot}6^N$ instances. The die is rendered in mild perspective, with the instruction sequence shown as a row of icons directly above it and the current instruction highlighted (\cref{fig:cache}). With the cache, the die is shown in its current pose after every step; without it, the die keeps its initial pose and turns to the final pose only in the last $20$ frames of the clip.

\paragraph{\textsc{Maze}.} The maze is a grid of $21{\times}21$ cells whose outer ring is wall, so that the ball moves within the inner $19{\times}19$ cells. It is generated by randomized depth-first search, which never reopens a visited cell, so the open cells form a spanning tree: the maze is perfect, with exactly one path between any two cells. The start and the $N$ waypoints are distinct open cells, drawn at random and well separated, and the waypoints are marked with their index. In the ground-truth video the ball follows the shortest path, found by breadth-first search, from the start to the first waypoint and from each waypoint to the next. With the cache, every cell the ball has visited is filled in.

\paragraph{\textsc{Gravity} and \textsc{Billiard}.} Both tasks are simulated in a $600{\times}600$ box and rendered at the resolution of each base model; lengths are given in simulation units. The ball has radius $R{=}18$ in both. Velocity and acceleration are defined per unit of simulated time, which is integrated in $8$ substeps in \textsc{Gravity} and $6$ in \textsc{Billiard}, independently of $N$. One interval is $48$ units of simulated time, so a clip at horizon $N$ covers $48N$ units and each frame advances $N/4$ units. In \textsc{Gravity} the ball is launched at speed $9.0$ under a gravity of $0.32$ and bounces with restitution $0.88$ off frictionless walls; in \textsc{Billiard} it moves at constant speed $4.25$ and reflects elastically. In both tasks the initial speed is the same for every instance and the initial direction is random. In \textsc{Gravity} the direction is uniform on $[0, 2\pi)$ and the initial position uniform on $[R{+}40,\, 600{-}R{-}40]$ along each axis, without rejection. The initial frame shows this direction as an arrow of fixed length, about $4R$, drawn from the center of the ball; it appears only in the initial frame, with and without the cache, since the initial direction is part of the problem. With the cache, the trail is the whole path of the ball from the initial frame to the current one, without fading; in \textsc{Billiard} it passes through the exact points of contact, so that it turns where the ball reflects.

The hexagon of \textsc{Billiard} is convex, with every interior angle at least $100^\circ$ and every edge at least $120$ long, about $3.3$ ball diameters. Its vertices lie at independent random angles around a center that is jittered by up to $20$ from the center of the box, at distances between $200$ and $295$ from it. The hexagon covers $40\%$ to $80\%$ of the box, every vertex is at least $8$ from the border of the box, and the ball starts at least $2R$ from every edge.

Unlike billiards with curved boundaries, billiards in polygons have zero Lyapunov exponents~[S1,\,S2]: a reflection off a flat edge does not amplify an error in direction, so a small error in the initial direction grows only linearly and the whole trajectory of \textsc{Billiard} can be compared with the ground truth. Only a vertex can separate nearby trajectories, so we resample any scene in which the ball, when it reflects off an edge, is within $0.25R$ of another edge, checked over twice the simulated time of $N{=}8$.

\begin{table}[h]
  \centering
  \small
  \caption{\textbf{Instance space and splits} per task and horizon $N$. Training, validation and test instances are disjoint, except for \textsc{Add}, \textsc{Mul} and \textsc{Dice} at $N{=}1$, whose instance spaces are too small to split and which are trained, validated and tested on all instances. Validation instances are used only to select the checkpoint.}
  \label{tab:supp_splits}
  \begin{tabular}{@{}llrrrr@{}}
    \toprule
    Task & & $N{=}1$ & $N{=}2$ & $N{=}4$ & $N{=}8$ \\
    \midrule
    \multirow{4}{*}{\textsc{Add}} & instances & 100 & $10^3$ & $10^5$ & $10^9$ \\
      & train & 100 & 700 & 3000 & 3000 \\
      & val   & 100$^\ast$ & 100 & 100 & 100 \\
      & test  & 100$^\ast$ & 200 & 500 & 500 \\
    \midrule
    \multirow{4}{*}{\textsc{Mul}} & instances & 81 & 810 & 81{,}000 & $8.1{\times}10^8$ \\
      & train & 81 & 510 & 3000 & 3000 \\
      & val   & 81$^\ast$ & 100 & 100 & 100 \\
      & test  & 81$^\ast$ & 200 & 500 & 500 \\
    \midrule
    \multirow{4}{*}{\textsc{Dice}} & instances & 144 & 864 & 31{,}104 & ${\sim}4{\times}10^7$ \\
      & train & 144 & 564 & 3000 & 3000 \\
      & val   & 144$^\ast$ & 100 & 100 & 100 \\
      & test  & 144$^\ast$ & 200 & 500 & 500 \\
    \midrule
    \multirow{3}{*}{Others$^\dagger$} & train & 3000 & 3000 & 3000 & 3000 \\
      & val   & 100 & 100 & 100 & 100 \\
      & test & 200 & 200 & 500 & 500 \\
    \bottomrule
  \end{tabular}
  \\[2pt]{\footnotesize $^\ast$All instances, identical to the training set. $^\dagger$\textsc{Maze}, \textsc{Gravity} and \textsc{Billiard}, each split separately; their instance spaces are continuous or effectively unbounded.}
\end{table}

\subsection{Scoring}
\label{sec:supp_metrics}

All scorers are rule-based: they read the state from the generated video with color thresholds, template matching and geometric rules, without a learned model or text recognition. They read only what is drawn with and without the cache---the answer line, the faces of the die, the ball---and never the cache itself, so the same scorer applies to all four schemes. Before scoring generated videos, each reader is run on the ground-truth renderings at every horizon $N\in\{1,2,4,8\}$, with and without the cache, and every scorer accepts all of them. On these renderings, the ball center located by the scorer in \textsc{Gravity} and \textsc{Billiard} is on average $0.89$ simulation units, about $0.05R$, from the true center, far below the tolerance of $2R$.

\paragraph{\textsc{Add} and \textsc{Mul}.} Only the answer line of the last frame is read. Its character positions are fixed by the layout, so each position is cropped and compared with templates of the digits, the plus sign and a blank, rendered in the same font; the template with the smallest mean squared difference in gray level over shifts of up to $3$ pixels is taken as the reading, which absorbs the jitter of generation. An instance succeeds if every position matches the correct result, including the positions that must stay blank. In \textsc{Mul} we read one position beyond the longest possible product, so a model that does not stop where the result ends fails.

\paragraph{\textsc{Dice}.} The camera is fixed, so the three visible faces of the die always project to the same three quadrilaterals in the frame. In each quadrilateral, shrunk to stay clear of the edges, dark pixels are grouped into connected components, and each component of plausible size counts as a pip. The three counts determine the pose of the die; an instance succeeds if, in the last frame, all three match those of the ground-truth pose.

\paragraph{\textsc{Maze}.} In every frame the ball is located from its pixels and mapped to a cell of the grid, which gives a sequence of cells. The waypoints are taken from the problem specification rather than from the frame. An instance succeeds if a single ball moves continuously, never enters a wall cell, and enters the cells of the waypoints in the prescribed order.

\paragraph{\textsc{Gravity} and \textsc{Billiard}.} The ground-truth trajectory is obtained by simulating from the initial position and direction given by the problem. In every frame the ball is located as the centroid of its pixels, converted to simulation coordinates so that the score does not depend on the rendering resolution. A frame is a hit if exactly one ball is detected and its center is within $2R$ of the ground-truth center in the same frame, and an instance succeeds if at least $80\%$ of the $T$ frames are hits (track@2R). Since positions are compared frame by frame, a ball that follows the right path at the wrong speed fails.

{\small
\section*{References for the appendix}
\begin{list}{}{\setlength{\leftmargin}{2.2em}\setlength{\labelwidth}{1.9em}\setlength{\labelsep}{0.3em}\setlength{\itemsep}{0pt}}
  \item[{[S1]}] Eugene Gutkin. Billiards in polygons: Survey of recent results. \emph{Journal of Statistical Physics}, 83:7--26, 1996.
  \item[{[S2]}] Serge Tabachnikov. \emph{Geometry and Billiards}. American Mathematical Society, 2005.
\end{list}
}

\end{document}